%% file: main.tex
\documentclass[conference]{IEEEtran}
\usepackage{times}

\usepackage[numbers]{natbib}
\usepackage{multicol}
\usepackage{hyperref}
\hypersetup{final=true}
\usepackage{url}
\usepackage{color}
\usepackage{amsmath}

\DeclareMathOperator*{\argmax}{arg\,max}
\usepackage{amssymb}
\usepackage{booktabs}
\usepackage{graphicx}
\usepackage{subfigure}
\usepackage{bbm}
\usepackage[normalem]{ulem}
\usepackage[labelfont=bf,font=small]{caption}

\renewcommand{\baselinestretch}{.96}

\begin{document}

\title{Enabling Robots to Communicate their Objectives}

\author{Sandy H. Huang$^\dagger$, David Held$^\dagger$, Pieter Abbeel$^\dagger$$^\ddagger$, Anca D. Dragan$^\dagger$ \\
$^\dagger$University of California, Berkeley, EECS  
$^\ddagger$OpenAI \\
}

\newcommand{\adnote}[1]{
 {\textcolor{blue}{\textbf{AD:#1}}}}

\newcommand{\shnote}[1]{
 {\textcolor{red}{\textbf{SH:#1}}}}
 
\newcommand{\eref}[1]{(\ref{#1})}
\newcommand{\sref}[1]{Sec. \ref{#1}}
\newcommand{\figref}[1]{Fig. \ref{#1}}
\newcommand{\prg}[1]{\noindent\textbf{#1. }} 
\newcommand{\be}[1]{\textbf{\emph{#1}}}
\newcommand\Tstrut{\rule{0pt}{2.6ex}}         
\newcommand\Bstrut{\rule[-0.9ex]{0pt}{0pt}}   

\maketitle

\input{abstract.tex}

\IEEEpeerreviewmaketitle

\input{introduction.tex}
\input{humanmodels.tex}
\input{exp_setup.tex}
\input{exp_sim.tex}
\input{exp_overall.tex}
\input{exp_utility.tex}

\input{exp_coverage.tex}
\input{discussion.tex}
\input{acknowledgments.tex}

\bibliographystyle{plainnat}
\bibliography{references}

\end{document}

%% file: abstract.tex
\begin{abstract}

The overarching goal of this work is to efficiently enable end-users to correctly anticipate a robot's behavior in novel situations. Since a robot's behavior is often a direct result of its underlying objective function, our insight is that end-users need to have an accurate mental model of this objective function in order to understand and predict what the robot will do.

While people naturally develop such a mental model over time through observing the robot act, this familiarization process may be lengthy.  Our approach reduces this time by having the robot model how people infer objectives from observed behavior,
and then it selects those behaviors that are maximally informative. 

The problem of computing a posterior over objectives from observed behavior is known as Inverse Reinforcement Learning (IRL), and has been applied to robots learning human objectives.
We consider the problem where the roles of human and robot are swapped. Our main contribution is to recognize that unlike robots, humans will not be \emph{exact} in their IRL inference.  We thus introduce two factors to define candidate approximate-inference models for human learning in this setting, and analyze them in a user study in the autonomous driving domain. We show that certain approximate-inference models lead to the robot generating example behaviors that better enable users to anticipate what it will do in novel situations.  Our results also suggest, however, that additional research is needed in modeling how humans extrapolate from examples of robot behavior.

\end{abstract}

%% file: introduction.tex
\section{Introduction}

Imagine riding in a self-driving car that needs to quickly change lanes to make a right turn. The car suddenly brakes in order to merge safely behind another  car, because it deems it unsafe to speed up and merge in front. A passenger who knows the car is defensive and that it values safety much more than efficiency would be able to anticipate this behavior. But passengers less familiar with the car would not anticipate this sudden braking, so they may be surprised and possibly frightened.

There are many reasons why it is beneficial for humans to be able to anticipate a robot's movements, from subjective comfort~\cite{Dragan_2014} to ease of coordination when working with and around the robot~\cite{Dragan_2015_7910,Sebanz_2006}. 
However, anticipation is challenging~\cite{Dragan_2014,Gielniak_2013,Matsui_2005}. Our goal is to enable end-users to accurately anticipate how a robot will act, even in \emph{novel situations that they have not seen the robot act in before}---like a new traffic scenario, or a new placement of objects on a table that the robot needs to clear. 

A robot's behavior in any situation is a direct consequence of the objective (or reward) function the robot is optimizing: (most) robots are rational agents, acting to maximize expected cumulative reward~\cite{Russell_2009}. Whether the robot's objective function is hard-coded or learned, it captures the trade-offs the robot makes between features relevant to the task. For instance, a car might trade off between features related to collision avoidance and efficiency~\cite{levine2012continuous}, with more ``aggressive'' cars prioritizing efficiency at the detriment of, say, distance to obstacles~\cite{sadigh2016information}.  

The insight underlying our approach is the following:

\begin{quote}
\emph{The key to end-users being able to anticipate what a robot will do in novel situations is having a good understanding of the robot's objective function. }
\end{quote}

\begin{figure}
    \centering
    \includegraphics[width=\columnwidth]{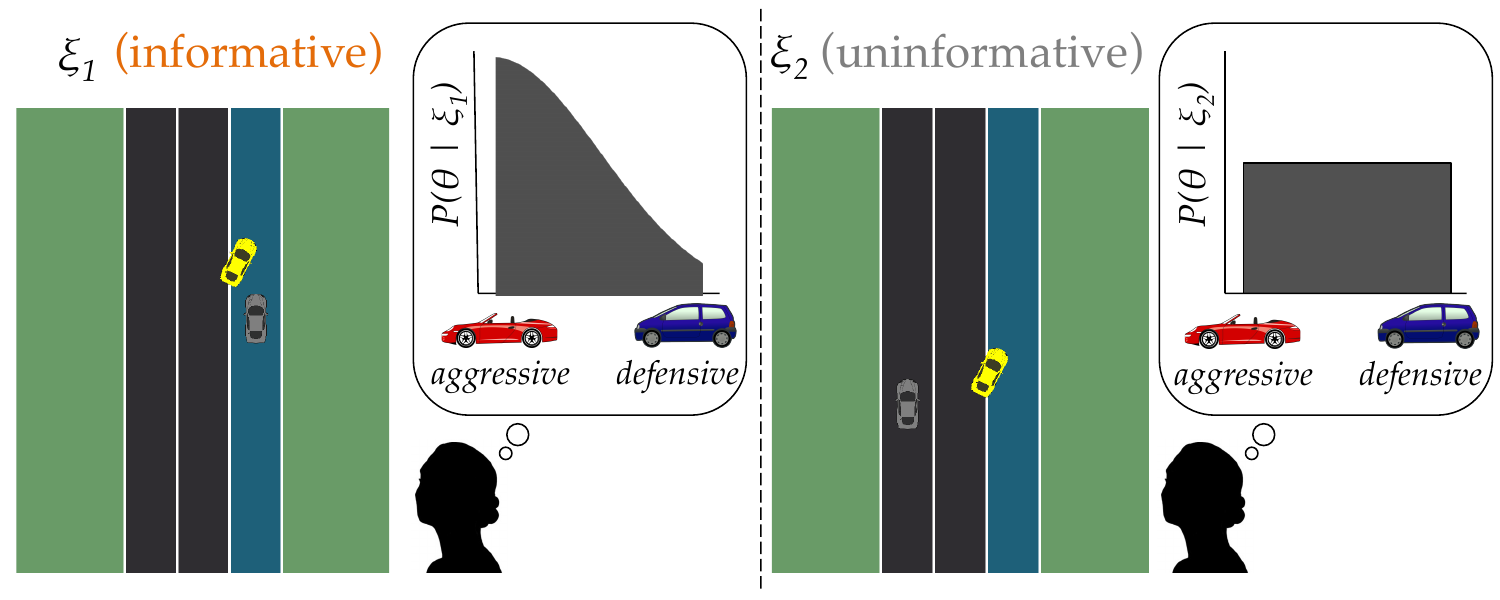}
    \caption{We show examples $\xi$ of the yellow autonomous car's behavior that are maximally informative in guiding the human toward understanding the robot's objective function (e.g., aggressive versus defensive). For instance, in environments where the car needs to merge into the right lane, its behavior is more informative when there is another car present (left) than when the lane is empty (right). We consider the case where the robot's objective function is represented by a linear combination of features, weighted by $\theta$.}
    \label{fig:front}
\end{figure}
Note that understanding the objective function does not mean users must be able to explicate it---to write down the equation, or even to assign the correct reward to a behavior or a state-action pair. Rather, users only need to have an implicit representation of what drives the robot's behavior, i.e., a qualitative understanding of the trade-offs the robot makes.

Fortunately, users will naturally improve their mental model of how a robot acts, given examples of the robot behaving optimally~\cite{Dragan_2014}. Further, evidence suggests that people will use this behavior to make inferences about the robot's underlying objective function~\cite{Baker_2009,Jara_2016}.

However, not all environments are equally informative. In many environments, a robot's optimal behavior does not fully describe the trade-offs that the robot would make in other environments, i.e. parts of the robot's objective will remain under-determined.
For example, an autonomous car driving down a highway with no cars nearby will drive at the speed limit and stay in its lane, regardless of its trade-off between  efficiency versus staying far away from other cars. Another example is when a car can change lanes without interacting with any other cars (\figref{fig:front}, right). An end-user mainly exposed to these types of behavior will have difficulty forming an accurate mental model of the robot's objective function and anticipating how the robot will behave in more complex scenarios.
On the other hand, suppose an autonomous car chooses to speed up and merge in front of another car, cutting it off (\figref{fig:front}, left). This scenario more clearly illustrates the trade-offs this car makes regarding safety versus efficiency. 

We focus on enabling robots to purposefully choose such \emph{informative} behaviors that actively communicate the robot's objective function. We envision a training phase for interaction, where the robot showcases informative behavior in order to quickly \emph{teach} the end-user what it is optimizing for.  

In order to choose the most informative example behaviors for communicating a robot's objective function to humans, we take an \emph{algorithmic teaching} approach~\cite{Balbach_2009,Goldman_1995,Koedinger_2013,Rafferty_2011,Zhu_2015,Zilles_2011}: we model how humans make inferences about the robot's objective function from examples of its optimal behavior, and use this model to generate examples that increase the probability of inferring the correct objective function. 

The opposite problem, machines inferring objective functions from observed human behavior, can be solved using Inverse Reinforcement Learning (IRL)~\cite{Ng_2000}. Prior work has investigated how to teach an objective function through example behavior to \emph{machine} learners running IRL~\cite{Cakmak_2012}. 
But the challenge in teaching \emph{people} instead of machines is that while machines can perform \emph{exact} inference, people are likely to be approximate in their inference.\footnote{Prior work has applied algorithmic teaching to teach humans, primarily for binary classification of images~\cite{Basu_2013,Bengio_2009,Khan_2011,Singla_2014}. In line with our work, Patil et al. show accounting for human limitations (in their case limiting the number of recalled examples)
improves teaching performance~\cite{Patil_2014}.} People do not have direct access to configuration-space trajectory and the exact environment state, whereas robots do, at least in kinesthetic teaching (and in~\cite{Cakmak_2012}).
People also cannot necessarily distinguish between a perfectly optimal trajectory for one objective and an ever-so-slightly suboptimal one~\cite{Vul_2014}.

Our main contribution is to introduce a systematic collection of approximate-inference models and, in a user study, compare their performance relative to the exact inference model. We focus on the autonomous driving domain, where a car chooses example behaviors that are informative about the trade-offs it makes in its objective function. We measure teaching performance---how useful the generated examples are in enabling users to anticipate the car's behavior in test situations---and find that one particular approximate-inference model significantly outperforms exact inference (the others perform on par). This supports our central hypothesis that accounting for approximations in user inference is indeed helpful, but suggests that we need to be careful about \emph{how} we model this approximate inference.

Further analysis shows teaching performance correlates with covering the full space of strategies that the robot is capable of adopting. For instance, the teaching algorithm cannot just show the car cutting people off; it also needs to show an example where it is optimal to brake and merge behind. We show the best results are obtained by a coverage-augmented algorithm that leverages an approximate-inference user model while encouraging full coverage of all possible driving strategies.

Our work takes a stab at an important yet under-explored problem of making robot objective functions more transparent to people.\footnote{Related work has explored communicating the payoff matrix in a collaborative (state-invariant) repeated game~\cite{Nikolaidis_2017}. Prior work on transparent robot behavior has explored explaining failure modes~\cite{Raman_2013,Tellex_2014}, verbalizing experiences~\cite{Perera_2016,Rosenthal_2016}, and explaining policies~\cite{Hayes_2017}.} This is important in the short term for human-robot interaction, as well as in the long term for building AI systems that are trustworthy and beneficial to people. Our results are encouraging, but also leave room for better models of how people extrapolate from observed robot behavior.  

%% file: humanmodels.tex
\section{Algorithmic Teaching of Objective Functions}
\label{sec:model}
We model how people infer a robot's objective function from its behavior, and leverage this model to generate informative examples of behavior.

\subsection{Preliminaries}
Let $S$ be the (continuous) set of states and $A$ be the (continuous) set of actions available to the robot. 
We assume the robot's objective (or reward) function is represented as a linear combination of features\footnote{We can make this assumption without loss of generality, as there are no restrictions on how complex these features can be.} weighted by some $\theta^*$~\cite{Abbeel_2004}:
\begin{equation}
R_{\theta^*}(s_t,a_t,s_{t+1};E) = \theta^{*\top} \phi(s_t,a_t,s_{t+1};E),
\end{equation}
where $s_t$ is the state at time $t$,  $a_t$ is the action taken at time $t$, and $E$ is the environment (or world) description. In the case of driving, $E$ contains information about the lanes, the trajectories of other cars, and the starting state of the robot. 

Given an environment $E$, the parameters $\theta$ of the objective function determine the robot's (optimal) trajectory $\xi^{\theta}_E$:
\begin{equation}
\xi^{\theta}_E = \argmax_{\xi_E \in \Xi_E} \theta^T \phi(\xi_E),
\end{equation}
where $\phi(\xi_E) = \sum_{t=0}^{T-1} \gamma^t \phi(s_t,a_t,s_{t+1}; E)$ and $\gamma$ is a discount factor between 0 and 1 that favors obtaining rewards earlier. $\Xi_E$ refers to all possible trajectories in environment $E$.

\subsection{Algorithmic Teaching Framework}
We model the human observer as starting with a prior $P(\theta)$ over what $\theta^*$ might be, and updating their belief as they observe the robot act. We assume the human knows the features $\phi(\cdot)$ relevant to the task.\footnote{In future work, we plan to study interactions for achieving common ground on what features are important.} The robot behaves optimally with respect to the objective induced by $\theta^*$, but as \figref{fig:front} shows, the details of the environment (e.g., locations of nearby cars and the robot's goal) influence the behavior, and therefore influence what effect the behavior has on the person's belief. 

To best leverage this effect, we search for a sequence of environments $E_{1:n}$ such that when the person observes the optimal trajectories in those environments, their updated belief places maximum probability on the correct $\theta$, i.e., $\theta^*$:
\begin{equation}
\argmax_{{E_{1:n}}} P(\theta^* | \xi_{E_{1:n}}^{\theta^*})
\end{equation}

To solve this optimization problem, the robot needs to model how examples update the person's belief, $P(\theta^* | \xi_{E_{1:n}}^{\theta^*})$.
We propose to model $P(\theta^* | \xi_{E_{1:n}}^{\theta^*})$ via Bayesian inference:
\begin{equation}
P(\theta | \xi_{E_{1:n}}^{\theta^*}) \propto P(\xi_{E_{1:n}}^{\theta^*} | \theta) P(\theta) = P(\theta) \prod_{i=1}^n P(\xi_{E_i}^{\theta^*} | \theta).\footnote{Conditional independence can be assumed, since $\theta$ contains all the information needed to calculate the probability of a trajectory. }
\label{eq:bayesmodel}
\end{equation}

With this assumption, modeling how people infer the objective function parameters reduces to modeling $P(\xi|\theta)$: how probable they would find trajectory $\xi$ if they assumed the robot optimizes the objective function induced by $\theta$. 
We explore different models of this, starting with exact-inference IRL as a special case. We then introduce models that account for the inexactness that is inevitable when real people make this inference.

\subsection{Exact-Inference IRL as a Special Case}
\label{sec:exactirl}

Inverse Reinforcement Learning (IRL)~\cite{Ng_2000} extracts an objective function from observed behavior by assuming that the observed behavior is optimizing some objective from a set of candidates. When that assumption is correct, IRL finds an objective function that assigns maximum reward (or minimum cost) to the observed behavior. 

Algorithmic teaching has been used with exact-inference IRL learners~\cite{Cakmak_2012}: the learner eliminates all objective functions which would \emph{not} assign maximum reward to the observed behavior. This can be expressed by the model in \eqref{eq:bayesmodel} via a particular distribution for $P(\xi_{E}^{\theta^*} | \theta) $:
\begin{equation}
P(\xi_{E}^{\theta^*} | \theta) = \begin{cases}
                        1, & \text{if $\forall \xi_{E}, \theta^\top \phi(\xi_{E}^{\theta^*} ) - \theta^\top \phi(\xi_{E}) \geq 0$}.\\
                        0, & \text{otherwise}.
                      \end{cases}
\end{equation}
This assumes people assign probability 0 to trajectories that are not perfectly optimal with respect to $\theta$, so those candidate $\theta$s receive a probability of zero.
Thus, each trajectory that the person observes completely eliminates from their belief any objective function that would not have produced exactly this trajectory when optimized. Assuming learners start with a uniform prior over objective functions, the resulting belief is a uniform distribution across the remaining candidate objective functions---$\theta$s for which all observed trajectories are optimal. 

While this is a natural starting point, it relies on people being able to perfectly evaluate whether a trajectory is \emph{the} (or one of the) \emph{global} optima of any candidate objective function. We relax this requirement in our approximate-inference models.

\subsection{Approximate-Inference Models}
\label{sec:approx_models}

We introduce a space of approximate-inference models, obtained by manipulating two factors in a 2--by--3 factorial design.

\prg{Deterministic versus Probabilistic Effect}
In the exact-inference model, a candidate $\theta$ is either out or still in: the trajectories observed so far have either shown that $\theta$ is impossible (because they were not global optima for the objective induced by that $\theta$), or have left it in the mix, assigning it equal probability as the other remaining $\theta$s.

We envision two ways to relax this assumption that a person can  identify whether a trajectory is optimal given a $\theta$:

One way is for observed trajectories to still either eliminate the $\theta$ or keep it in the running, but to be more conservative about which $\theta$s get eliminated. That is, even if the observed trajectory is not a global optimum for a $\theta$, the person will not eliminate that $\theta$ if the trajectory is \emph{close enough} (under some distance metric) to the global optimum. We call this the \emph{deterministic} effect.

A second way is for observed trajectories to have a \emph{probabilistic} effect on $\theta$s: rather than eliminating them completely, trajectories can make a $\theta$ less likely, depending on how far away its optimal trajectory is from the observed trajectory.

In both cases, $P(\xi_E^{\theta^*} | \theta)$ no longer depends on the example trajectory being optimal with respect to $\theta$. Instead, it depends on the \emph{distance} $d(\cdot,\cdot)$ between $\xi_E^{\theta}$, the optimal trajectory for $\theta$, and $\xi_E^{\theta^*}$, the observed trajectory which is optimal given $\theta^*$. 

Given some distance metric $d$ and hyperparameters $\tau, \lambda > 0$,
\begin{itemize}
\setlength\itemsep{0.5em}
\item For deterministic effect, \\
$P(\xi_E^{\theta^*} | \theta) \propto 0$ if $d(\xi_E^{\theta},\xi_E^{\theta^*}) > \tau$, or $1$ otherwise.
\item For probabilistic effect, \\
$P(\xi_E^{\theta^*} | \theta) \propto e^{-\lambda \cdot d(\xi_E^{\theta},\xi_E^{\theta^*})}$.\footnote{We noticed normalizing this distribution produced very similar results to leaving it unnormalized, so we do the latter in our experiments, analogous to other algorithmic teaching work not based on reward functions \cite{Singla_2014}.}
\end{itemize}

\begin{figure*}[t!]
\centering
\includegraphics[width=\textwidth]{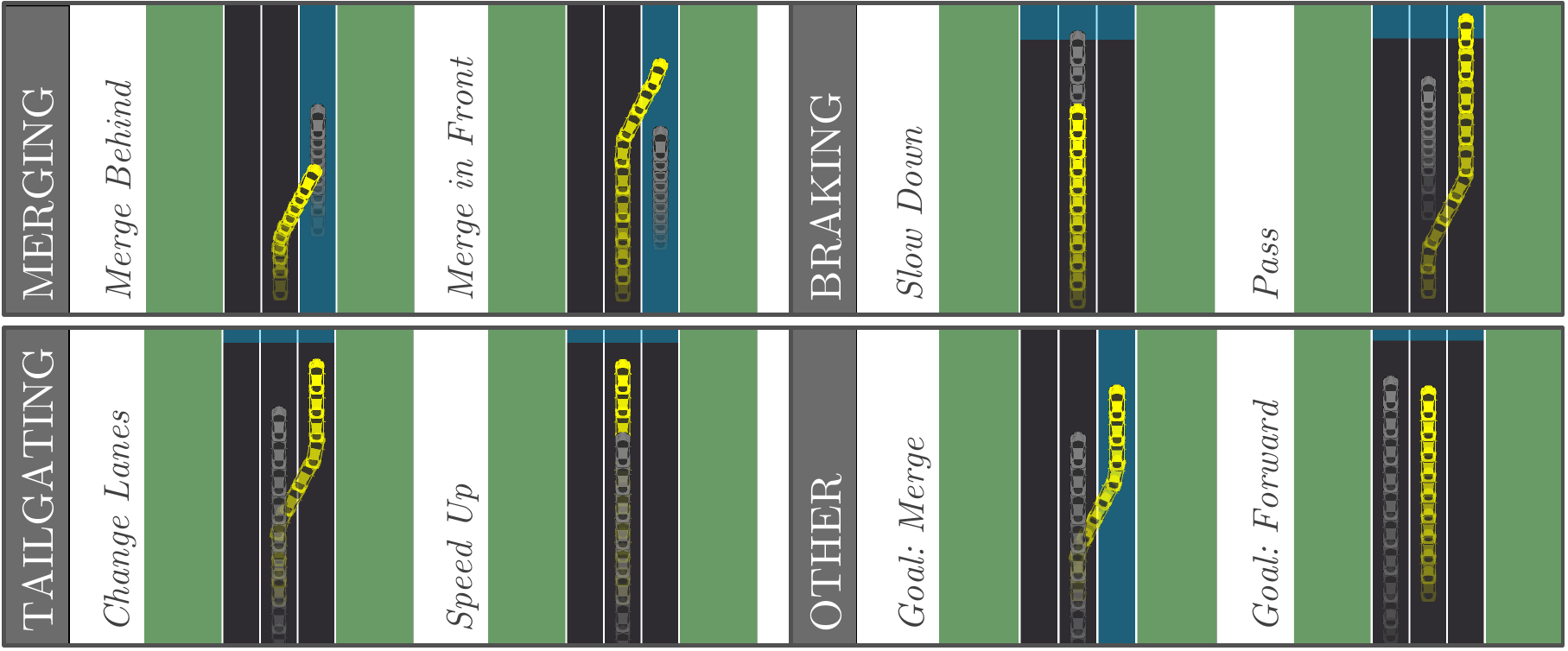}
\caption{The possible driving environments cluster naturally into four classes, with two trajectory strategies per class. Each image shows the trajectories of the autonomous car (yellow) and non-autonomous car (gray) in a particular environment. Positions later in the trajectory are more opaque. The goal of the autonomous car in each environment is highlighted in blue: merge into the right lane or drive forward.}
\label{fig:env_classes}
\end{figure*}

The deterministic effect results in conservative hypothesis elimination: it models a user who will either completely eliminate a $\theta$ or not, but who will not eliminate $\theta$s with optimal trajectories close to the observed trajectory. In contrast, the probabilistic effect decreases the probability of $\theta$s with far away optimal trajectories, never fully eliminating any. 

The exact-inference IRL model (\sref{sec:exactirl}) is a special case with deterministic effect and a reward-based distance metric with $\tau = 0$; it assumes there is no approximate inference.

\prg{Distance Metrics}
\label{sec:distance}
Both deterministic and probabilistic effects rely on the person's notion of how close the optimal trajectory with respect to a candidate $\theta$ is from the observed trajectory. We envision that closeness can be measured either in terms of the reward of the trajectories with respect to $\theta$, or in terms of the trajectories themselves. 

We explore three options for $d$. The first depends on the reward.
This distance metric models people with difficulty comparing the cumulative discounted rewards of two trajectories, with respect to a given setting of the reward parameters. So, if in environment $E$ the observed trajectory $\xi_E^{\theta^*}$ has almost the same reward as $\xi_E^{\theta}$, the optimal trajectory with respect to $\theta$, then $P(\xi_E^{\theta^*} | \theta)$ will be high.

\begin{itemize}
\item reward-based\footnote{Note that this is always positive because $\xi_E^{\theta}$ has maximal reward w.r.t. $\theta$.}: $d(\xi_E^{\theta},\xi_E^{\theta^*})=\theta^\top \phi(\xi_E^{\theta}) - \theta^\top \phi(\xi_E^{\theta^*}).$
\end{itemize}

The second option depends not on reward, but on the physical trajectories. It assumes it is not high reward that can confuse people about whether the observed trajectory is optimal with respect to $\theta$, but rather physical proximity to the true optimal trajectory: this models people who cannot perfectly distinguish between perceptually-similar trajectories.

\begin{itemize}
\item Euclidean-based: $d(\xi_E^{\theta},\xi_E^{\theta^*}) = \frac{1}{T}\sum_{t=1}^T||s_{E,t}^{\theta} - s_{E,t}^{\theta^*}||_2$
where $s_{E,t}^{\theta}$ is the state at time $t$ for the trajectory $\xi_E^{\theta}$.\footnote{This requires an appropriate representation of the state space, e.g., if the dimensions of $s_{E,t}^{\theta}$ have different ranges, normalization may be necessary.}
\end{itemize}

Finally, a  more conservative version of the Euclidean distance metric is the strategy-based metric. The idea here is that for any environment $E$, trajectories generated by candidate $\theta$s can be clustered into types, or strategies. 
The strategy-based metric assumes people do not distinguish among trajectories that follow the same strategy.  For instance, people will consider all trajectories in which the robot speeds up and merges in front of another car to be equivalent, and all trajectories in which the robot merges behind the car to be equivalent. So, if in environment $E$ the observed trajectory and the optimal trajectory with respect to $\theta$ have the same strategy, then $P(\xi_E^{\theta^*} | \theta) \propto 1$.

\begin{itemize}
\item strategy-based: $d(\xi_E^{\theta},\xi_E^{\theta^*}) = 0$ if $\xi_E^{\theta}$ and $\xi_E^{\theta^*}$ are
in the same trajectory strategy cluster, $\infty$ otherwise.
\end{itemize}

\prg{Relation to MaxEnt IRL} MaxEnt IRL~\cite{Ziebart_2008} is an IRL algorithm that assumes demonstrations are noisy (i.e., not necessarily optimal). In our setting, we instead assume demonstrations are optimal but the learner is approximate. These two sources of noise result in the same model: the MaxEnt distribution is equivalent to our \emph{probabilistic reward-based} model:
\begin{align}
P(\xi_E^{\theta^*}|\theta) &\propto e^{\lambda\theta^T\phi(\xi_E^{\theta^*})} \\
& \propto e^{\lambda(\theta^T\phi(\xi_E^{\theta^*}) - \theta^T\phi(\xi_E^{\theta}))} = e^{-\lambda \cdot d_{r}(\xi_E^{\theta},\xi_E^{\theta^*})}.
\end{align}

\subsection{(Submodular) Example Selection} Given a learner model $\mathcal{M}$ that predicts $P_{\mathcal{M}}(\theta^*|\xi^{\theta^*}_{E_{1:n}})$, our approach greedily selects environment $E_t$ to maximize $P_{\mathcal{M}}(\theta^*|\xi^{\theta^*}_{E_{1:t}})$, which is estimated by uniformly sampling candidate $\theta$s. We allow the model to select up to ten examples; it stops early if no additional example improves this probability.

This greedy approach is near-optimal for deterministic effect with a uniform prior, since in this case maximizing $P_{\mathcal{M}}(\theta^*|\xi^{\theta^*}_{E_{1:t}})$ is equivalent to maximizing $-\sum_{\theta \neq \theta^*} P_{\mathcal{M}}(\theta|\xi^{\theta^*}_{E_{1:t}})$, which is a non-decreasing monotonic submodular function~\cite{Nemhauser_1978}. This function is non-decreasing because adding example trajectories $\xi^{\theta^*}_{E}$ can only eliminate candidate $\theta$'s, not add them, and we assume the set of candidate $\theta$s considered by the human does not change over time. Additionally, a particular observed trajectory $\xi^{\theta^*}_E$ eliminates the same set of $\theta$s no matter when it is added to the sequence. Thus, showing that example later on in the sequence cannot eliminate more $\theta$s than adding it earlier, which makes this function submodular.

\subsection{Hyperparameter Selection} We would like to select values for hyperparameters $\tau$ and $\lambda$ (for deterministic and probabilistic effect, respectively) that accurately model human learning in this domain. $\tau$ and $\lambda$ affect the informativeness of examples. If $\tau$ is too large, then most environments will be uninformative, since the observed trajectory will be within $\tau$ distance away from optimal trajectories of many $\theta$s, so those $\theta$s will not be eliminated. Thus, $P_{\mathcal{M}}(\theta^*|\xi^{\theta^*}_{E_{1:n}})$ will be low. On the other hand, if $\tau$ is too small, then some environments will be extremely informative, so only one or a few examples will be selected before no further improvement in $P_{\mathcal{M}}(\theta^*|\xi^{\theta^*}_{E_{1:n}})$ can be achieved. Analogous reasoning holds for $\lambda$.

We expect humans to be teachable (i.e., $\tau$ cannot be too large) and to have approximate rather than exact inference (i.e., $\tau$ cannot be too small), so they would benefit from observing several examples rather than just one or two. Based on this, we select $\tau$ and $\lambda$ for each approximate-inference model by choosing the value in $\{10^{-5}, 10^{-4}, \ldots, 10^4, 10^5\}$ that results in an increase from $P_{\mathcal{M}}(\theta^*|\xi^{\theta^*}_{E_{1}})$ to $P_{\mathcal{M}}(\theta^*|\xi^{\theta^*}_{E_{1:n}})$ of at least 0.1, and selects the largest number of unique examples to show.

%% file: exp_setup.tex
\section{Example Domain}
We evaluate how our proposed approximate-inference models perform for teaching the
driving style of a simulated autonomous car. In this domain, participants
witness examples (in simulation) of how the car drives, with the goal of being able to anticipate how it will drive when they ride in it. 

\prg{Driving Simulator} We model the dynamics of the car with the bicycle vehicle model~\cite{Taheri_1990}. Let the state of the car be $\mathbf{x} = [x \enspace y \enspace \theta \enspace v \enspace \alpha]^\top$, where $(x,y)$ are the coordinates of the center of the car's rear axle, $\theta$ is the heading of the car, $v$ is its velocity, and $\alpha$ is the steering angle. Let $\mathbf{u} = [u_1 \enspace u_2]^\top$ represent the control input, where $u_1$ is the change in steering angle and $u_2$ is the acceleration. Additionally, let $L$ be the distance between the front and rear axles of the car. Then the dynamics model of the vehicle is
\begin{equation}
[\dot{x} \enspace \dot{y} \enspace \dot{\theta} \enspace \dot{v} \enspace \dot{\alpha}] = [v \cdot cos(\theta) \enspace\enspace v \cdot sin(\theta) \enspace\enspace \frac{v}{L} tan(\alpha) \enspace\enspace v \cdot u_1 \enspace\enspace u_2]
\end{equation}

\prg{Environments} We consider a total of 
21,216 environments of highway driving configurations (Table~\ref{tab:env_options}). Each environment has three lanes and a single non-autonomous car. The autonomous car always starts in the middle lane with the same initial velocity, whereas the initial location and velocity of the single non-autonomous car varies.

\begin{table}[!t]
\centering
\caption{Environment Parameters}
\label{tab:env_options}
\begin{tabular}{ll}
\hline
\textbf{Axis of Variation} & \textbf{Acceptable Values} \Tstrut\Bstrut \\ \hline
Goal & \begin{tabular}[c]{@{}l@{}}{[}merge to right, drive forward{]} \Tstrut \Bstrut \\ \end{tabular} \Tstrut\Bstrut \\ \hline
\begin{tabular}[c]{@{}l@{}}Distance between autonomous\\ and non-autonomous car\end{tabular} & \begin{tabular}[c]{@{}l@{}}{[}-240, -220,\ldots, -100{]} \Tstrut \\ {[}100, 120, \ldots, 240{]}\end{tabular} \Tstrut\Bstrut\\ \hline
Lane of non-autonomous car & {[}Left, Center, Right{]} \Tstrut \Bstrut \\ \hline
\begin{tabular}[c]{@{}l@{}}Initial velocity, 
non-autonomous \end{tabular} & {[}20, 25, \ldots, 80{]} \Tstrut \Bstrut\\ \hline
\begin{tabular}[c]{@{}l@{}}Acceleration time, non-autonomous \end{tabular} & {[}0, 0.5, 1, 1.5, 2{]} \Tstrut \Bstrut\\ \hline
\begin{tabular}[c]{@{}l@{}}Final velocity, non-autonomous \Tstrut \\ car (if acceleration time $\neq$ 0)\end{tabular} & {[}20, 30, 70, 80{]} \Bstrut\\ \hline
\end{tabular}
\end{table}

These driving environments naturally fall into four classes, with two trajectory strategies per class (\figref{fig:env_classes}):

\emph{Merging}: when the non-autonomous car starts in the right lane, and the goal in this environment is to merge into the right lane.
The two trajectory strategies are to either speed up and merge ahead of the non-autonomous car, or slow down and merge behind
the non-autonomous car.

\emph{Braking}: when the non-autonomous car starts in the center lane in front of the autonomous car, and the goal  is to drive forward. The two trajectory strategies are to either keep driving in the center lane behind the non-autonomous car, or merge into another lane to pass it.

\emph{Tailgating}: when the non-autonomous car starts in the center lane behind the autonomous car, and the goal is to drive forward. The two trajectory strategies are to either change lanes to avoid the tailgater, or speed up to maintain a safe distance from the tailgater.

\emph{Other}: all environments not included in one of the first three. The autonomous car is able to reach its goal without any interaction with the non-autonomous car.

\prg{Reward Features} We use the following reward features $\phi(\xi)$:

\emph{distance to other car}: $\sum_{t=1}^T \gamma^t e^{-\frac{1}{2} (p_t - p'_t)^\top \Sigma_t^{-1} (p_t - p'_t)}$; each term corresponds to a multivariate Gaussian kernel, where $p_t = [x_t, y_t]^\top$, $p'_t$ is the non-autonomous car's position, and $\Sigma_t$ is chosen so that the major axis is along the non-autonomous car's heading.

\emph{acceleration}, squared: $\sum_{t=1}^{T-1} \gamma^t (v_{t+1} - v_t)^2$

\emph{deviation from initial speed}, squared: $\sum_{t=1}^T \gamma^t (v_t - v_1)^2$

\emph{turning}: $\sum_{t=1}^T \gamma^t |\theta_t - \theta_1|$

\emph{distance from goal}: $\sum_{t=1}^T \gamma^t \max(0, (x_1+w) - x_t)^2$ if the goal is to merge into the right lane, and $y_T$ if the goal is to drive forward. $w$ is the width of one lane.

The last four features do not depend on the environment, so we normalize such that the maximum value of each feature across all trajectories is 1 and the minimum is 0. We use $\gamma = 1$.

\prg{Optimal $\theta$} We select $\theta^* = [-64 \enspace -0.1 \enspace -1 \enspace -0.1 \enspace -0.5]^\top$, a reward function that is not overly cautious about staying away from other cars.

%% file: exp_sim.tex
\section{Analysis of Approximate\\ Inference Models with Ideal Users}
\label{sec:simulated}

\begin{figure}
\centering
\includegraphics[width=\columnwidth]{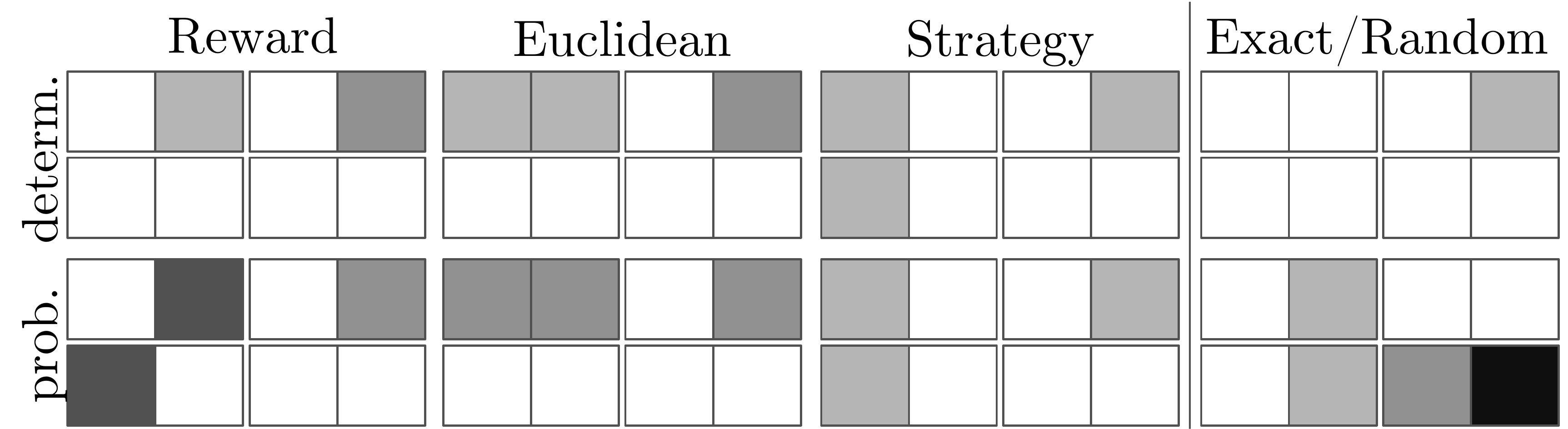}
\caption{The number of examples shown in each of the eight trajectory classes for the approximate-inference models, exact inference model, and random baseline. White = 0 examples shown, and black = 4. The environment classes are arranged in the 2x4 grid as in \figref{fig:env_classes}.}
\label{fig:examples_shown}
\end{figure}
\begin{figure}[t!]
\centering
\includegraphics[width=0.8\columnwidth]{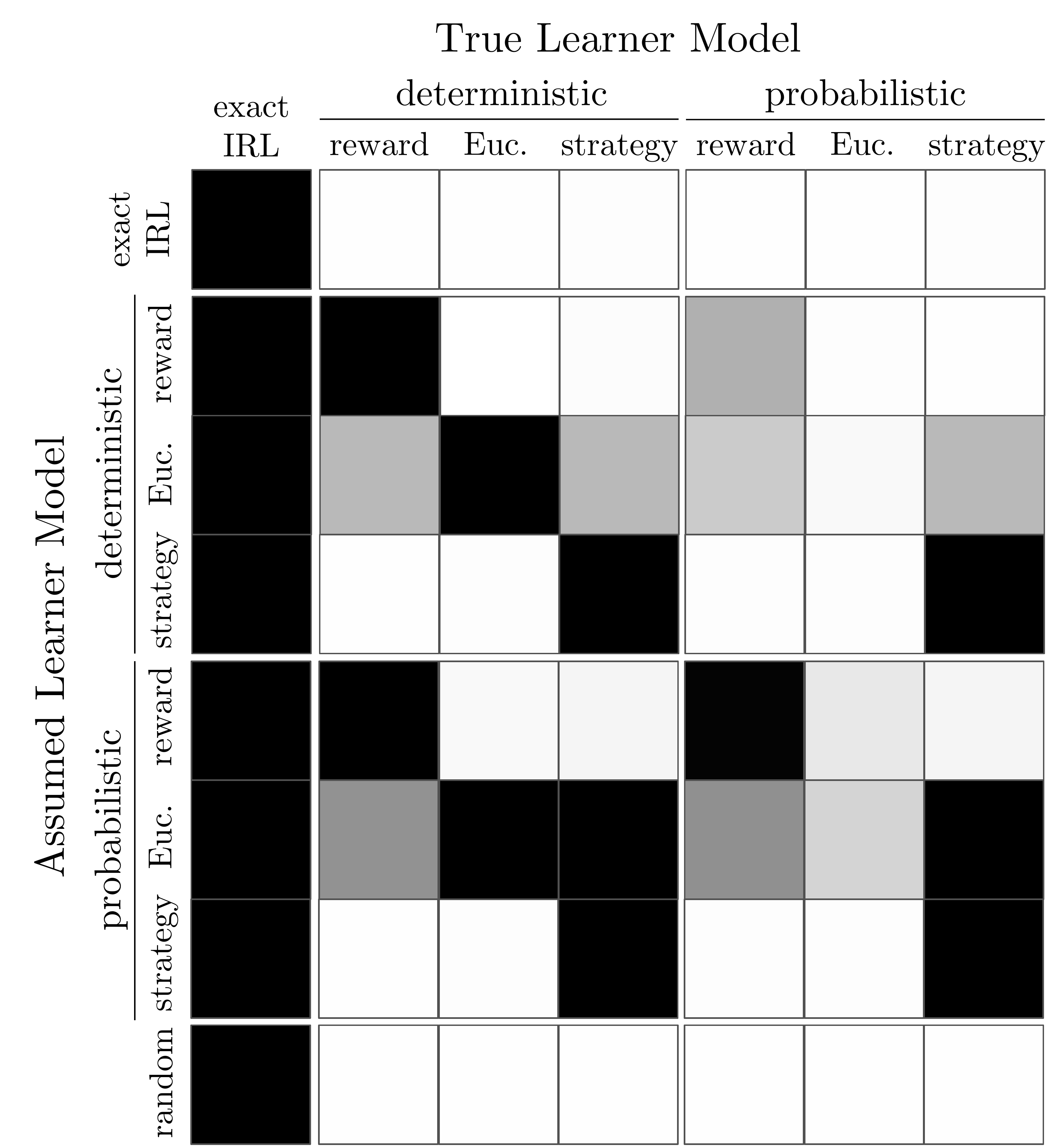}
\caption{The performance of each user model as evaluated by all other models. White indicates $P_{\mathcal{M}}(\theta^*|\xi^{\theta^*}_{E_{1:n}}) = 0$, where $\mathcal{M}$ is the true learner model and environments $E_{1:n}$ are chosen based on the assumed learner model. Black indicates $P_{\mathcal{M}}(\theta^*|\xi^{\theta^*}_{E_{1:n}}) = 1$.}
\label{fig:sim_users}
\end{figure}

In \sref{sec:approx_models}, we introduced six possible approximate-inference user models $\mathcal{M}$. They all model people as judging candidate $\theta$s based on the distance between the trajectory they observed and the optimal trajectory with respect to $\theta$, but they differ in what the distance metric is, and whether they completely eliminate candidate $\theta$s (deterministic effect) or smoothly re-weight them (probabilistic effect).

Here, we investigate how well algorithmic teaching with these models performs for teaching $\theta^*$ to ideal users. First, we generate a sequence of examples for each of our approximate-inference models $\mathcal{M}$, by greedily maximizing $P_{\mathcal{M}}(\theta^*|\xi^{\theta^*}_{E_{1:n}})$. We also generate the sequence for the exact-inference model and include a random sequence, for a total of eight sequences.

\prg{Types of Examples Selected} \figref{fig:examples_shown} summarizes the types of examples that each algorithm selected for its optimal sequence. The exact-inference model selects a single example, because that is enough to completely eliminate all other $\theta$s. This works well for an ideal user running exact inference, but our hypothesis is that it does not work as well for real users.

\prg{Relative Evaluation} We evaluate algorithmic teaching on seven ideal users, whose learning exactly matches one of our six approximate-inference models or exact-inference IRL. We measure, for each ``user'' $M$,  the probability they assign to the correct objective function parameters, $P_{\mathcal{M}}(\theta^*|\xi^{\theta^*}_{E_{1:n}})$, given $\xi^{\theta^*}_{E_{1:n}}$ from each of the eight generated sequences. 

\figref{fig:sim_users} shows the results. First, we see for any ideal user $M$, the sequence generated by assuming a learner model $M$ performs best at teaching that user. This is by design---that sequence of examples is optimized to teach $M$.

Looking across the columns of \figref{fig:sim_users}, we see all eight sequences perform equally well for teaching an exact IRL learner (column 1)---even random, because it provides enough examples to perfectly eliminate all incorrect $\theta$s. This suggests exact IRL does not accurately model real users, whose performance likely varies based on which examples they see. Looking across the rows, we notice assuming a Euclidean distance approximation when generating examples (rows 3 and 6) leads to robust performance across different user models.
\sref{sec:user} evaluates these generated sequences on real users.

Finally, the random sequence is very uninformative for all ideal users except exact IRL, showing the utility of algorithmic teaching. We explore this utility with real users in \sref{sec:utility}.

%% file: exp_overall.tex
\section{User Study}
\label{sec:user}
We now evaluate whether approximate-inference models are useful with real, as opposed to ideal, users.

\subsection{Experiment Design}
\label{exp_design1}

\prg{Manipulated Variables} 
We manipulate whether algorithmic teaching assumes exact-inference or approximate-inference.
For the approximate-inference case, we manipulated two variables:
the effect of approximate inference (either \be{deterministic} or \be{probabilistic}) and the distance metric (\be{reward-based},
\be{Euclidean-based}, or \be{strategy-based}), in a 2--by--3 factorial design, for a total
of six approximate inference models. For the strategy-based distance metric, the type of effect
does not matter since distances are either $0$ or $\infty$, so there are five unique
approximate-inference models.

We show the participant one training environment at a time,
in the order that the examples were selected by each algorithm. 

\prg{Dependent Measures} In the end, we are interested in how well human participants
learn a specific setting of reward parameters $\theta^*$ from the training examples.
Since we cannot ask them to write down a $\theta$, or to drive according to how they think the car will drive (people can drive like themselves, but not so easily like others),
we evaluate this by testing each participant's ability to 
identify the trajectory produced by $\theta^*$ in a few test environments.
For each test environment, we show the participant four trajectories and ask them
to select the one that most closely matches the autonomous car's driving style, and report their confidence (from 1 to 7) for how closely each of the four trajectories matches the driving style.

We have two dependent variables: whether participants correctly identify
$\xi^{\theta^*}_{E_{\text{test}}}$ for each test environment
$E_{\text{test}}$, and their confidence in selecting that trajectory.
We combine the two in a confidence score: the confidence if they 
are correct, negative of the confidence if they are not---this
score captures that if one is incorrect, it is better to be not confident about it.

We use rejection sampling to select test environments in which there are a
wider variety of possible robot trajectories. To make sure the four trajectories do not look too similar,
we ensure the rewards of alternate trajectories under $\theta^*$ are below a certain threshold.
In order to not bias the measure, we select one test environment for each
of the two trajectory strategy clusters
in each of the three informative environment classes, for a total of six test
environments. For each test environment, we show
two trajectory options in each strategy cluster.

\prg{Hypothesis} \emph{Accounting for approximate inference significantly improves performance (the confidence score).}
We leave open which approximate-inference models work well and which do not, since the goal is to identify which captures users' inferences the best. 

\prg{Subject Allocation}
We used a between-subjects design, since examples of the same reward function interfere with each other. 
We ran this experiment on a total of 191 participants across the six conditions, recruited via Amazon Mechanical Turk.
At the end of the experiment, we ask participants what the two possible goals were, to filter out those who were not paying attention.
30 out of 191 (15.7\%) answered incorrectly.
The average age of the 161 non-filtered participants was 37.0 ($SD=11.0$). The gender ratio was 0.46 female.

\subsection{Analysis}
\prg{Number of Examples} Different algorithms produce different numbers of examples. Exact-inference IRL might produce as few as one example (and does in our case). Approximate-inference models produce more, and random can produce an almost unlimited amount if allowed. Thus, a possible confound in our experiment is the number of examples. 

\begin{figure}[t!]
\centering
\includegraphics[width=\columnwidth]{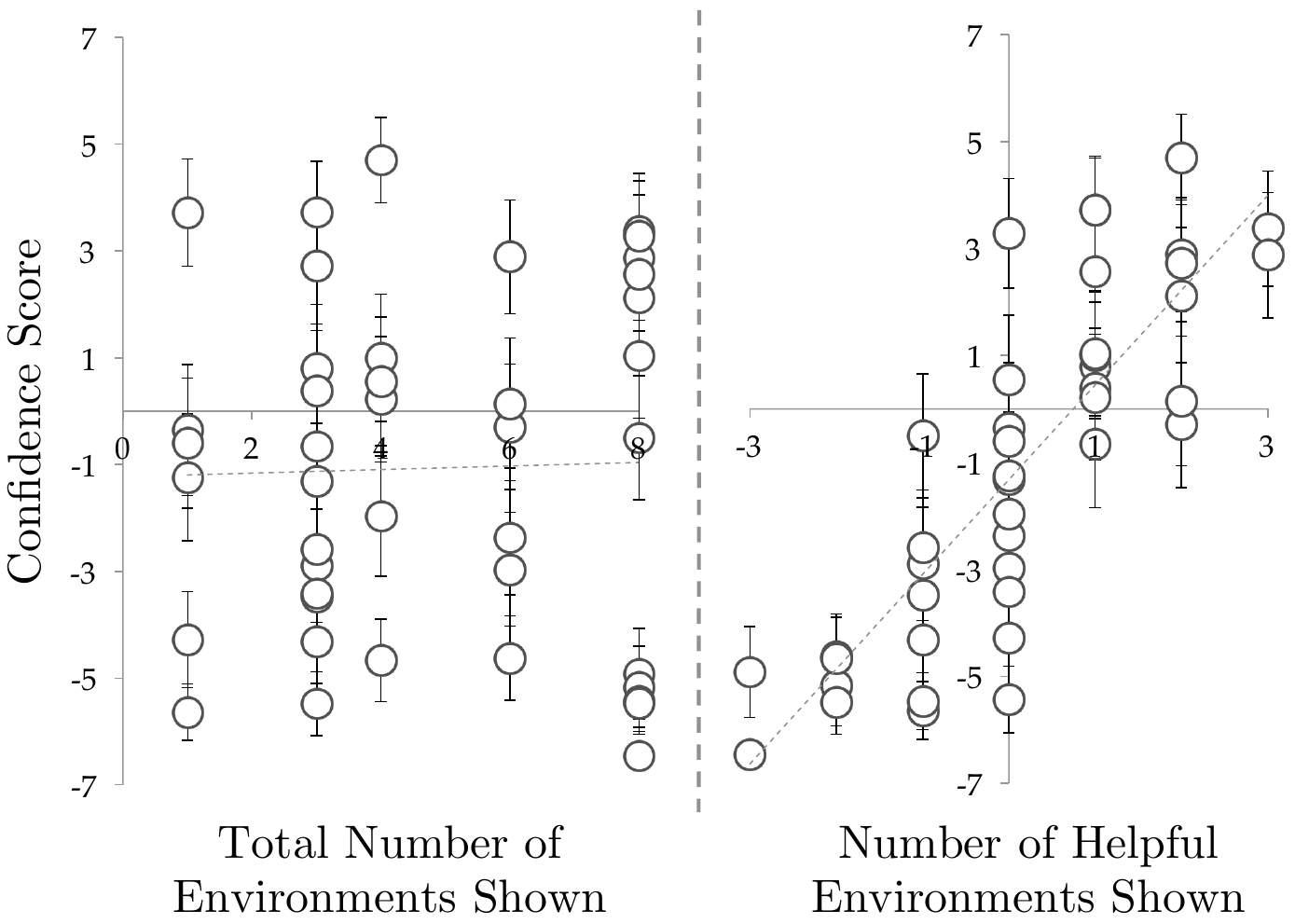}
\caption{\textbf{Left:} Lack of correlation between total number of environments shown in a condition, and average participant performance on each test example in that condition. \textbf{Right:} Correlation exists between the number of helpful training examples shown for an environment class, and average participant performance on the test example corresponding to that environment class.}
\label{fig:corr}
\end{figure}

We checked whether this is indeed a confound: do more examples help? Surprisingly, we found no correlation between the number of examples and performance: the sample Pearson correlation coefficient is $r = 0.03$ 
(\figref{fig:corr}, left). This suggests that example quantity matters less than example quality.

\begin{figure}[t!]
\centering
\includegraphics[width=\columnwidth]{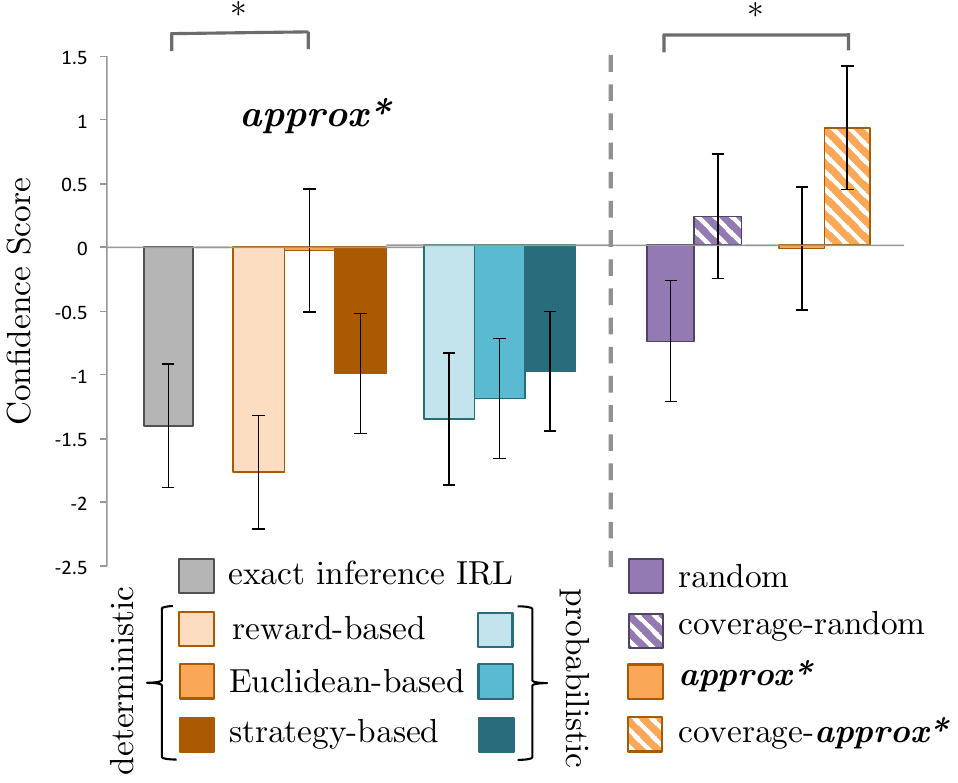}
\caption{Performance of human participants on identifying the autonomous car's trajectory in test environments, after seeing the example trajectories selected by the approximate-inference models (left) and after adding coverage (right) to the sequence of environments selected by the best-performing approximate-inference model, \be{approx*}. Participants in the coverage-\be{approx*} condition performed significantly better than those in the random condition. In contrast, enforcing coverage but selecting random sequences (coverage-random) does not lead to statistically significantly better performance.}
\label{fig:noisemodels_sig}
\end{figure}

\prg{Approximate-Inference Models} We begin our analysis by comparing the different approximate-inference models. We ran a factorial ANOVA on \be{confidence score} with distance measure and determinism as factors (\figref{fig:noisemodels_sig}, left). We found a marginal effect for \be{distance} ($F(2,163)=2.69$, $p=.07$), with Euclidean-based distance performing the best (as suggested by our simulation experiment in \sref{sec:simulated}, where Euclidean was the most robust across different ideal users), and reward-based distance performing the worst. 

Euclidean-based might be better than reward-based because people decide to keep imperfect $\theta$s not when the trajectory they see obtains high reward under that $\theta$, but when it is visually similar to what optimizing for $\theta$ would have produced. Euclidean-based might be better than strategy-based because people differentiate between trajectories even when they follow the same strategy. For example, a trajectory that gets very close to another car would be in the same strategy class as a trajectory that stays farther away, as long as they both merge behind the other car, but these two trajectories may give people very different impressions of the car's driving style.

There was no effect for determinism. On average, probabilistic models performed ever-so-slightly worse than deterministic ones, and the difference was largest for Euclidean distance. This might be because keeping track of what is possible is easier than maintaining an entire probability distribution.

The best approximate-inference model used the Euclidean-based distance with deterministic effects. We refer to this as the \be{approx*} model (\figref{fig:noisemodels_sig}, left).

\prg{Central Hypothesis: Utility of Approximate Inference} Despite showing more examples, most approximate-inference models did not perform much better than exact-inference IRL.
This shows that not just any approximate-inference model is useful. However, it does not imply that \emph{no} approximate-inference model is useful. 
To test the utility of accounting for approximate inference, we compared the best model, \be{approx*}, with exact-inference IRL, and found a significant improvement (Welch's t-test $p = 0.025$). 

\begin{quote}
\emph{This supports our central hypothesis, that accounting for approximate inference in our model of human inferences about objective functions helps, with the caveat that not just any approximation will work.}
\end{quote}

%% file: exp_utility.tex
\section{Utility of Algorithmic Teaching}
\label{sec:utility}

So far, we have tested our central hypothesis, that accounting for approximate IRL inferences can indeed improve the performance of algorithmic teaching of humans. 
While this is promising, we also want to test the utility of algorithmic teaching itself: whether our approach is preferable not just to algorithmic teaching with exact inference, but to the robot not actively teaching. Instead, the person must learn from the robot's behavior in environments that it happens to encounter.

\subsection{Baselining Performance}

\prg{Baseline Condition} We ran a follow-up study comparing algorithmic teaching with a sequence of optimal trajectories in \emph{random} environments---simulating that the robot does not choose these, but instead happens to encounter them.  We recruited 33 users for this condition. The average age of the 28 non-filtered participants was 33.3 ($SD = 9.6$). The gender ratio was $0.46$ female.

\prg{Controlling for Confounds}
 There are several variables that could confound this study.
 First, when generating the random sequence, we might get very lucky or unlucky and 
 generate a particularly informative or uninformative one. To avoid this, 
 we randomly sample 1000 random sequences with the desired number
 of examples, and sort them based on $P_{\mathcal{M}}(\theta^*|\xi^{\theta^*}_{E_{1:n}})$
 where $\mathcal{M}$ is the exact-inference IRL model. Then we choose the median
 sequence in that ranking as our random sequence, which will have median informativeness.

 Second, different algorithms produce different numbers of examples.
 For instance, exact-inference IRL only selects one example, which
 eliminates all $\theta$s other than $\theta^*$---because in that environment the
 optimal trajectories for all $\theta$s are at least slightly
 different than that for $\theta^*$.
 To give the random baseline the best chance, we choose to select eight environments for it, which is the maximum number of examples shown by any of the other conditions. Since the majority
 of environments are uninformative (i.e., not in the merging, braking, or tailgating classes),
 providing the random condition with eight environments is needed to not put it at a serious disadvantage.

\prg{Analysis}
Algorithmic teaching with our approximate inference model did outperform the random baseline, albeit not significantly (Welch's t-test $p = 0.23$ when comparing participants' confidence scores). Algorithmic teaching \emph{without} accounting for approximate inference actually seems to perform poorly compared to random (\figref{fig:noisemodels_sig}, right).

\prg{Coverage}
Digging deeper, we realized users tended to perform well on test cases for strategies in which they had seen a training example. In addition, for each pair of environment strategies $A$ and $B$ (e.g., merge-in-front and merge-behind), if users did not see an example from strategy $A$ but saw one from strategy $B$, their performance was worse than if they did not see any examples from either $A$ or $B$! In other words, if users see one trajectory strategy in the training examples and not the opposite strategy, they tend to think the autonomous car will always take the first strategy in that environment type.

Based on this observation, given that $x$ training examples are shown in environment class $A$ and $y$ from $B$, we define the number of helpful environments shown in $A$ as equal to $x$ if $x > 0$, and equal to $-y$ otherwise. We found a strong correlation between the number of ``helpful'' environments shown and users' confidence scores, with a Pearson correlation coefficient of $r = 0.83$ ($p = 1.4\times10^{-11}$) (\figref{fig:corr}, right).

We leverage this result to introduce augmented algorithms that ensure coverage of strategies.

%% file: exp_coverage.tex
\subsection{Coverage-Augmented Algorithmic Teaching}

Since coverage correlates with better user performance, we 
add a coverage term to our optimization over trajectories
$\xi_{E_{1:n}}^{\theta*}$:
\begin{equation}
\argmax_{\xi_{E_{1:n}}^{\theta*}} P_{\mathcal{M}}(\theta^* | \xi_{E_{1:n}}^{\theta*}) + \lambda \sum_c \mathbbm{1} [ \exists i, h(\xi_{E_{i}}^{\theta*}) = c ],
\end{equation}
where the sum is over trajectory strategy clusters $c$ and the function $h$
maps a trajectory to the strategy it belongs to.

We set $\lambda_{t} = \mathbbm{1}
[P_{\mathcal{M}}(\theta^* | \xi_{E_{1:t}}^{\theta*}) - P_{\mathcal{M}}(\theta^* | \xi_{E_{1:t-1}}^{\theta*}) < \epsilon]$,
so that 
only after no examples will significantly increase the probability of $\theta^*$,
extra examples are selected to provide coverage across the strategies.
We select these extra examples by choosing the best with
respect to the approximate-inference model $\mathcal{M}$, to ensure they are informative.
Using this approach, we augment our best approximate-inference model, \be{approx*}, to
achieve coverage. 

 \subsection{User Study on Coverage}

We next run a study to test the benefit of coverage. 

\prg{Manipulated Variables} We manipulate two variables: whether we augmented
the training examples with coverage, and whether
we used a user model to generate the examples or sampled uniformly.
We select our best model for the former, \be{approx*}. From the
previous experiment, we have obtained user performance data
along the no-coverage dimension---for random and \be{approx*}---so
we run this experiment on only the two new conditions that incorporate coverage: coverage-random and coverage-\be{approx*}.
We generate random sequences with coverage by randomly selecting
exactly one random environment
from each of the eight trajectory strategy classes.

\prg{Dependent Measures} We keep the same dependent measures as in our previous
Mechanical Turk experiment (\sref{exp_design1}).

\prg{Hypothesis} We hypothesize coverage augmentation
improves user performance in both conditions, random and \be{approx*},
compared to the respective conditions without coverage.

\prg{Subject Allocation} We ran this experiment between-subjects, on a total of 63 participants across
the two conditions.
The average age of the 53 non-filtered participants was 34.43 ($SD = 9.0$). The gender ratio was 0.53 female.

\prg{Analysis}
We ran a factorial ANOVA on \be{confidence score} with coverage and model as factors. We found a marginal effect for coverage ($F(1,107)=1.82$, $p=.07$), suggesting that coverage improves performance. There was no interaction effect, suggesting that coverage helps regardless of using a user model for teaching or not.

Coverage-\be{approx*} performed best out of the four conditions. The coverage augmentation enabled it to significantly outperform the random baseline (with a Welch's t-test $p = 0.049$), which suggests coverage is useful. Coverage augmentation did not enable the coverage-random condition to outperform the random baseline ($p = 0.159$), which suggests the approximate-inference model is useful (\figref{fig:noisemodels_sig}, right).

Overall, coverage alone helped, but was not sufficient to outperform the baseline. From the previous experiment, we know that the approximate-inference user model helped, but was also not sufficient to outperform the baseline. The improvement is largest (and significant, modulo compensating for multiple hypotheses) when we have a coverage-augmented approximate-inference IRL model.

 \begin{quote}
 \emph{When leveraged together, coverage with the right approximate-inference model have a significant teaching advantage over random teaching, as well as over IRL models that assume exact-inference users. }
 \end{quote}

%% file: discussion.tex
\section{Discussion}

\prg{Summary} We take a step towards communicating robot objective functions to people. We found that an approximate-inference model using a deterministic
Euclidean-based update on the space of candidate objective function parameters performed best
at teaching real users, and outperforms algorithmic teaching that assumes exact inference. 
We additionally found after augmenting such a model with a coverage objective, it outperformed
letting the user passively familiarize to the robot. 

\prg{Limitations and Future Work} Our results reveal the promise of algorithmic \emph{teaching} of robot objective functions. However, the coverage results suggest that an IRL-only model is not sufficient for capturing how people extrapolate from observed robot behavior. 
People may also depend on direct policy learning techniques and/or infer an objective function based on a more rich set of features. Alternatively, people may use hierarchical reasoning, in which they first determine which trajectory strategy the robot will take (for which they need examples of each possible strategy), and then select the most likely trajectory within that strategy cluster.

Furthermore, in this work we focused on the robot's physical behavior as a communication channel because people naturally infer utility functions from it. Future work could augment this with other channels, like visualizations of the objective function or language-based explanations.

%% file: acknowledgments.tex
\section{Acknowledgments}
This research was funded in part by Air Force \#16RT0676, Intel Labs, an NSF
CAREER award (\#1351028), and the Berkeley Deep Drive
consortium. Sandy Huang was supported by an NSF Fellowship.